\documentclass[10pt,letter]{article}

\input{preface.tex}
\arxivtrue
\usepackage{amsmath}
\usepackage{amssymb}
\usepackage{authblk}
\usepackage[margin=1in]{geometry}

\begin{document}

\title{
AGATHA:
Automatic
Graph-mining
And
Transformer based
Hypothesis generation
Approach
}

%Following KDD conference tradition, reviews are not double-blind, and author names and affiliations should be listed.

\author[1]{
  Justin Sybrandt
  \thanks{\href{mailto:jsybran@clemson.edu}{jsybran@clemson.edu}}
}
\author[1]{
  Ilya Tyagin
  \thanks{\href{mailto:ityagin@clemson.edu}{ityagin@clemson.edu}}
}
\author[2]{
  Michael Shtutman
  \thanks{\href{mailto:shtutmanm@sccp.sc.edu}{shtutmanm@sccp.sc.edu}}
}
\author[1]{
  Ilya Safro
  \thanks{\href{mailto:isafro@clemson.edu}{isafro@clemson.edu}}
}
\affil[1]{
  School of Computing,
  Clemson University
}
\affil[2]{
  Drug Discovery and Biomedical Sciences,
  University of South Carolina
}

\date{}

\maketitle

\begin{abstract}
  Medical research is risky and expensive. Drug discovery, as an example,
  requires that researchers efficiently winnow thousands of potential targets to a small
  candidate set for more thorough evaluation. However, research groups
  spend significant time and money to perform the experiments necessary to determine
  this candidate set long before seeing intermediate results.  Hypothesis
  generation systems address this challenge by mining the wealth of publicly
  available scientific information to predict plausible research
  directions.  We present \pymoliere{}, a deep-learning hypothesis generation
  system that can introduce data-driven insights earlier in
  the discovery process.  Through a learned ranking criteria, this
  system quickly prioritizes plausible term-pairs among entity sets, allowing us to
  recommend new research directions.
  We massively validate our system with a
  temporal holdout wherein we predict connections first introduced after 2015 using data published beforehand.
  We additionally explore biomedical sub-domains, and demonstrate
  \pymoliere's predictive capacity across the twenty most popular relationship
  types.  This system achieves best-in-class performance on an established
  benchmark, and demonstrates high recommendation scores across subdomains.
\ifthesis
  Additionally, \pymoliere{} is fast and scalable, is constructed using
  distributed data preparation and training, and can analyze thousands of
  hypotheses per-minute.
\fi
  {\bf Reproducibility:} All code, experimental data, and
  pre-trained models are available online: \pymoliererepo{}.
\end{abstract}

\ifarxiv
\else


\fi

\section{Introduction}
\label{sec:introduction}

As the rate of global scientific output continues to climb~\cite{van2014global},
an increasing portion of the biomedical discovery process is becoming a ``big
data'' problem. For instance, the US National Library of Medicine's (NLM)
database of biomedical abstracts, \emph{MEDLINE}, has steadily increased the
number of papers added per-year, and has added significantly over 800,000 papers
every year since 2015~\cite{nlm2018citations}. This wealth of scientific
knowledge comes with the overhead cost payed by practitioners who often
struggle to keep up with the state-of-the-art.
% First on cutting block
\ifthesis
In 2018 alone, there was an
average of 103 papers added to MEDLINE per \textit{hour}, or one new paper every
\textit{35 seconds}.
\fi

Buried within the large and growing MEDLINE database are many undiscovered
implicit connections --- those relationships that are implicitly discoverable,
yet have not been identified by the research community. One connection of this
type was first proposed and subsequently discovered by Swanson and Smalheiser in
the mid-to-late 1980's~\cite{swanson1986undiscovered}. Their landmark finding,
using only the co-occurrences of keywords across MEDLINE titles, was to
establish a connection between fish oil and Raynaud's
Syndrome~\cite{swanson1986fish}. At that time, it was known that fish oil
modified various bodily properties, such as blood viscosity, which were key
factors pertaining to Raynaud's syndrome. However, while each explicit
relationship was known, the \textit{implicit} relationship was not discovered
before Swanson's ARROWSMITH hypothesis generation system identified the
connection algorithmically.

Modern advances in machine learning, specifically in the realms of text and
graph mining, enable contemporary hypothesis generation systems to identify
fruitful new research directions while taking far more than title co-occurrence
rates into account. 
Modern systems predict missing links on domain specific graphs, such as BioGraph on gene-disease network~\cite{liekens2011biograph} or MeTeOR on the term-co-occurrence graph~\cite{wilson2018automated}. Other systems focus on identifying relevant key terms, similar to Swanson's work, but using modern techniques. For instance, Jha et al. study the evolution of word embedding spaces over time to learn contemporary trends relevant to particular queries~\cite{jha2018concepts}. Further work by Jha et al. continues to study the joint evolution of corpora and ontologies within biomedical research~\cite{jha2019hypothesis}. Another approach is to produce visualizations for interpretation by domain scientists~\cite{spangler2015accelerating}, such as the closed-source Watson for Drug Discovery~\cite{chen2016ibm}.
Moliere, our prior hypothesis generation system~\cite{sybrandt2017}, produces data for scientific interpretation in the form of LDA topic models~\cite{blei2012probabilistic}. Additional work produced heuristically-backed ranking criteria to help automate the analysis process~\cite{sybrandt2018a}.

While prior hypothesis generation systems have been valuable in real-world explorations, such as Swanson's fish-old and Raynaud's syndom finding~\cite{swanson1986fish}, Watson's discovery of ALS treatments~\cite{chen2016ibm}, or Moliere's discovery of DDX3 inhibition as a treatment for HIV-associated neurodegenerative disease~\cite{bakkar2018artificial}, there remains significant drawbacks to the state of the art.
Most systems require significant human oversight to produce useful results~\cite{spangler2014automated,chen2016ibm,sybrandt2017}, or are only tested on very small evaluation sets~\cite{jha2018concepts,jha2019hypothesis,gopalakrishnan2016generating,sang2018gredel}. Systems still using the ``ABC'' model of discovery~\cite{jha2018concepts,jha2019hypothesis,kim2019context}, posed by Swanson in 1986~\cite{swanson1986fish}, face many known limitations such as reduced scalabiltiy and a bias towards incremental discoveries~\cite{smalheiser2012literature}.

To overcome these limitations, we present a new hypothesis generation system that scales to the entirety of biomedical literature, and is backed by efficient deep-learning techniques to enable thousands of queries a minute, enabling new types of queries.
This system constructs a new semantic multi-layered
graph, and places its millions of nodes into a shared embedding. From there, we
use a \textit{transformer encoder}
architecture~\cite{vaswani2017attention} to learn a ranking criteria between
regions of our semantic graph and the plausibility of new research connections.
Because our graph spans all of MEDLINE, we are able to generate hypotheses from
a large range of biomedical subdomains.  \emph{Other than our prior work~\cite{sybrandt2017}, we are
unaware of any system that is capable of the same breadth of cross-domain discovery that is also open source, or even just publicly available for comparison.}
Because we efficiently pre-process our graph and its
embeddings, we can perform hundreds of queries per-second on GPU, which enables
new many-to-many recommendation queries that were not previously feasible.
Because we replace our heuristically determined ranking criteria from our prior
work~\cite{sybrandt2018a} with a learned ranking criteria, we achieve
significantly improved performance, as demonstrated by an increase in benchmark
performance using the same training and validation from and ROC AUC of
0.718~\cite{sybrandt2018b} to 0.901.

\ifthesis
To expand the interpretabiliy of our proposed system's output, we additionally
provide an updated version of our prior topic-modeling approach to operate
efficiently on the much larger sentence-focused semantic graph. While these
topic-model queries do take longer (a few minutes, compared to a fraction of a
second), they can provide descriptive results for biomedical scientists. Because
our topic-model query process selects relevant sentences, as opposed to the
prior model that selected whole abstracts, we observe that our resulting topic
models are more descriptive regarding the query at hand. We envision that these
descriptive queries can supplement the automated discovery enabled by the
deep-learning model.
\fi

\noindent{\bf Our contribution:} 

\noindent(1) We introduce a novel  approach to construct large semantic graphs
that use the granularity of sentences to represent documents. These graphs are
constructed using a pipeline of state of the art NLP techniques that have been
customized  for understanding scientific text, including
SciBERT~\cite{beltagy2019scibert} and ScispaCy~\cite{neumann2019scispacy}.

\noindent(2) We deploy our deep-learning transformer-based model that trained to
predict likely connections between term-pairs at scale.  This is done by
embedding our proposed semantic graph to encode all sentences, entities,
n-grams, lemmas, UMLS terms, MeSH terms, chemical identifiers, and SemRep
predicates~\cite{arnold2015semrep} in a common space using the PyTorch-BigGraph
embedding~\cite{lerer2019biggraph}.

\noindent(3) We validate our system using the massive
validation techniques presented in~\cite{sybrandt2018a}, and also demonstrate
the ability of \pymoliere{} to generalize across biomedical subdomains. For
instance, in the scope of ``Gene - Cell Function'' relationships, our system
has a top-10 average precision of 0.83, and a mean-reciprocal-rank of 0.61.

This system is open-source, easily installed, and all prepared data
and trained models are available to perform hypothesis queries at \pymoliererepo{}.

\section{Background}
\label{sec:background}

\ifthesis
Our proposed system, \pymoliere{}, is a novel deep-learning approach to
hypothesis generation, enabled by recent advances in text- and graph-mining
techniques. This section expounds our conceptual influences and incorporated
technologies.
\fi

\noindent{\bf Hypothesis Generation Systems.} Swanson posited that
\textit{undiscovered public knowledge}, those facts that are implicitly
available but not explicitly known, would accelerate scientific discovery if an
automated system were capable of returning them~\cite{swanson1986undiscovered}.
His work established what is now known as the ``A-B-C'' model of
literature-based discovery~\cite{smalheiser2017rediscovering}. This formulation
follow that a hypothesis generation system, given two terms $A$ and $C$, should
uncover some likely $B$-terms that explain the quality of a potential $A-B-C$
connection. This technique fueled Swanson's own system,
ARROWSMITH~\cite{swanson1986fish}, and still forms the backbone of some
contemporary successors~\cite{kim2019context}.

\ifthesis
The ABC model has significant limitations. Firstly, many real-world scientific
hypotheses cannot be so easily distilled into a single set of ``first-order''
interactions. Instead, many connections may be better described as longer
$A-B-C-\dots$ connection paths or more complicated structures. Secondly, any system that returns only a set of
$B$-terms will be limited to small-scale searches unless it also provides an
automatic way to quantify connection plausibility. Otherwise, biomedical
researchers will spend a significant amount of valuable time studying query
results, rather than performing necessary experiments.
\fi

Our former approach to address these challenges is posed by the Moliere
system~\cite{sybrandt2017}, and its accompanying plausibility ranking
criteria~\cite{sybrandt2018a}.  This system expands on the $A-B-C$ model by
describing a range of connection patterns, as represented by an LDA topic
model~\cite{blei2012probabilistic}, when receiving an $A, C$ query.  To do so,
the Moliere system first finds a short-path of interactions bridging the $A-C$
connection from within a large semantic graph. This structure includes nodes
that correspond to different entity types that are both textual and biomedical,
such as abstracts, predicate statements, genes, diseases, proteins, etc.  Edges
between entities indicate similarity. For instance, an edge may exist between an
abstract and all genes discussed within it, or between two proteins that are
discussed in similar contexts.  Using the short-path discovered within the
semantic network between $A$ and $C$, the Moliere system also reports an LDA
topic model~\cite{blei2012probabilistic}.  This model summarizes popular areas
of conversation pertaining to abstracts identified near to the returned path.
As a result, the user can view various fuzzy clusters of entities and the
importance of interesting concepts across documents.

To reduce the burden of topic-model analysis on biomedical researchers, the
Moliere system is augmented by a range of techniques that automatically quantify
the plausibility of the query based on its resulting topic models. Our measures,
such as the embedding-based similarity between keywords and topics, as well as
network analytic measures based on the topic-nearest-neighbors network, were
heuristically backed, and were combined into a meta-measure to best understand
potential hypotheses. Using this technique, we both validated the overall
performance of the Moliere system, and used it to identify a new gene-treatment
target for HIV-associated neurodegenerative disease through the inhibition of
DDX3X~\cite{aksenova2019inhibition}.

\ifthesis
The work presented here departs from heuristically-backed prior necessities. Our
\pymoliere{} semantic graph is built using sentences, not abstracts, as the
primary node type, and uses ScispaCy~\cite{neumann2019scispacy} in order to
produce higher-quality graph content pertaining to key terms and entities. The
resulting graph, with a different overall schema, is then embedded by the
PyTorch-BigGraph heterogeneous graph embedding
technique~\cite{lerer2019biggraph}. Now, instead of performing expensive
short-path queries, generating topic models, or applying heuristically-backed
measures, we formulate knowledge discovery as a deep-learning problem, and learn
to rank fruitful new research directions directly from the data using a
combination of graph embeddings and a transformer-encoder
network~\cite{vaswani2017attention}.
\fi

\noindent{\bf Related and Incorporated Technologies.}
\ifthesis
In order to prepare the wealth of biomedical information stored in Medline
abstracts for deep-learning queries, we leverage a range of software tools and
machine learning techniques.
\fi

\noindent{\bf SemRep}~\cite{arnold2015semrep} is a utility that extracts
\textit{predicate statements} in the form of ``subject-verb-object'' from the
entirety of Medline. This utility further classifies its predicate components
into the set coded keywords provided by the Unified Medical Language System
(UMLS), and a small set of coded verb-types. These UMLS terms provide a way to
unify synonyms and acronyms from across medicine. Additionally, all content
extracted by SemRep is provided in the Semantic Medical Database (SemMedDB) for
direct use.

\ifthesis
\noindent{\bf Dask}~\cite{rocklin2015dask} is a library for writing distributed
data-processing pipelines in Python. As a result of using this library,
\pymoliere{} data preparation is efficiently completed across a large
cluster of workers. Furthermore, this library allows us to implement modular
functional components of the processing pipeline, enabling easier extensions.
\fi

\noindent{\bf ScispaCy}~\cite{neumann2019scispacy}, a version of the popular
spaCy text processing library provided by AllenNLP, is designed to properly
handle scientific text. Using a deep-learning approach for its part-of-speech
tagging, dependency parsing, and entity recognition, this tool achieves
state-of-the-art performance on a range of scientific and biomedical linguistic
benchmarks. Additionally, this software is optimized sufficiently to operate on
each sentence of MEDLINE, which numbers over 188 million as of 2020.

\noindent{\bf SciBert}~\cite{beltagy2019scibert} is a version of the BERT
transformer model for scientific language.
This model learns representations for each word part in a given
sentence.
\ifthesis
Word parts are derived from the WordPiece
algorithm~\cite{wu2016google} when trained on a sample of scientific full-text
papers.
\fi
The resulting embeddings for each word part are determined by its
relationship to all other word parts.  As a result, the output word-part
embeddings are highly content-dependent, and homographs, words with the same
spelling but different meanings, receive significantly different
representations. 
\ifthesis
We use this model to learn embeddings per-sentence that capture
scientific content. These embeddings inform the sentence-nearest-neighbors
network component of our semantic graph.
\fi

\noindent{\bf FAISS}~\cite{johnson2017faiss}, the open-source similarity-search
utility, is capable of computing an approximate
nearest-neighbors network for huge point clouds.
\ifthesis
We adapt this tool for use
within Dask in order to compute the nearest neighbors edges between the
SciBert embeddings for all sentences in MEDLINE.
\fi
This technique scales to
various graph sizes by its modular component set, and we choose PQ-quantization
and $k$-means bucketing to reduce the dimensionality of our sentences, and
reduce the search space per-query. 

\noindent{\bf PyTorch-BigGraph} (PTBG)~\cite{lerer2019biggraph} is an
open-source, large-scale, distributed graph-embedding technique 
aimed at heterogeneous information networks~\cite{shi2016survey}. These
graphs consist of nodes of various types, connected by typed edges. We define
each node and relationship type contained in our semantic graph
as input to this embedding technique.
PTBG distributes edges such that all machines compute on
disjoint node-sets. We choose to
encode edges through the dot product of transformed embeddings, which we explain
in more detail in Section~\ref{sec:methods}.
\ifthesis
Using the
Hogwild!~\cite{recht2011hogwild} optimization technique, distributed workers are
unrestricted by locking while performing this optimization, which has an added
regularization effect.
\fi

\noindent{\bf The Transformer}~\cite{vaswani2017attention} model
is built with multi-headed attention. Conceptually, this 
mechanism works by learning weighted averages per-element of the input sequence,
over the entire input sequence.  Specifically, this includes three projections
of each element's embedding, represented as packed matrices: $Q$, $K$, and $V$.
\ifthesis
Each projection functions differently, with $Q$ acting as a ``query'' that is
compared against ``keys'' $K$ and ``values'' $V$.
\fi
The specific mechanism is
defined as follows, with $d_k$ representing the dimensionality of each $Q$ and
$K$ embedding:
\begin{equation}
  \text{Attention}(Q, K, V) =
  \text{softmax}\left(\frac{QK^\intercal}{\sqrt{d_k}}\right)V
\end{equation}

The ``multi-headed'' aspect of the transformer indicates that the attention
mechanism is applied multiple times per-layer, and recombined to form a joint
representation. If $W^{(x)}$ indicates a matrix of learned weights, then this
operation is defined as:
\begin{equation}
\begin{aligned}
  \text{MultiHead}(X) &= [h_1;\dots;h_k]W^{(4)} \\
  \text{where } h_i &=
  \text{Attention}\left(XW_i^{(1)},XW_i^{(2)},XW_i^{(3)}\right)
\end{aligned}
\end{equation}

\ifthesis
While the transformer model was initial proposed for sequence-to-sequence
modeling, and includes both an ``encoder'' and ``decoder'' stack of
attention layers,
we note that the self-attention layer fundamentally
performs a \textit{set} operation. In fact, text models such as
BERT~\cite{devlin2018bert} require the addition of a positional encoding to each
input token to ensure that positional information is not erased by
self-attention. 
\fi
By using only the encoder half of the transformer model, and
by omitting any positional mask or encoding, we apply the self-attention
mechanism to understand input sets while reducing the effect of the arbitrary
ordering imposed by a sequence model. One encoder layer is defined as:
\begin{equation}
\begin{aligned}
  \mathcal{E}(X) &= \text{LayerNorm}(FF(\alpha) + \alpha) \\
  \text{where } FF(\alpha) &= \text{max}\left(0, \alpha W^{(5)}\right)W^{(6)} \\
  \text{and }\alpha &= \text{LayerNorm}(\text{MultiHead}(X) + X)
\end{aligned}
\end{equation}

\ifthesis
By composing multiple $\mathcal{E}$ encoders, we create the full encoder stack.
While some order-sensitive operations do exist within the encoder stack, such as
the operation that merges multiple attention heads into a joint feed-forward
layer, we observe these artifacts are overcome during the training process by
randomizing the order of input elements.
\fi
\section{Data Preparation}
\label{sec:methods}

\begin{figure*}
    \centering
\ifarxiv
    \includegraphics[width=0.90\linewidth]{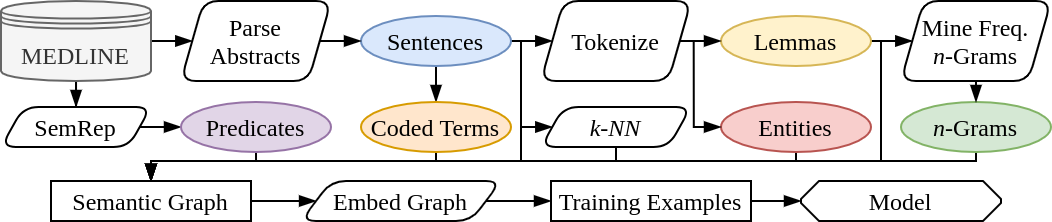}
\else
    \includegraphics[width=0.70\linewidth]{images/system_diagram.png}
\fi
    \caption{System Diagram of the \pymoliere{} process.
    }
    \label{fig:system_diagram}
\end{figure*}

\begin{figure}
    \centering
\ifarxiv
    \includegraphics[width=0.4\linewidth]{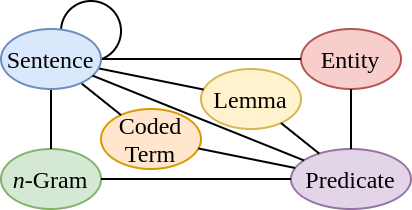}
\else
    \includegraphics[width=0.6\linewidth]{images/graph_schema.png}
\fi
    \caption{\pymoliere{} multi-layered graph schema. %Each bubble represents a class of nodes, while each line represents a category of edges.
    }
    \label{fig:graph_schema}
\end{figure}

\begin{figure}
    \centering
\ifarxiv
    \includegraphics[width=0.3\linewidth]{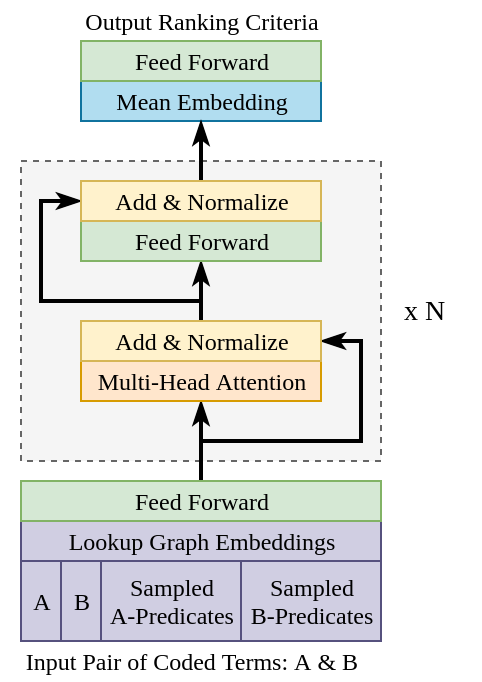}
\else
    \includegraphics[width=0.7\linewidth]{images/hypothesis_prediction_model.png}
\fi
    \caption{
      \label{fig:hypothesis_prediction_model}
      \pymoliere{} ranking transformer encoder. 
      \ifthesis
      Given entity-pair and neighborhoods, looks up graph embeddings and produce ranking criteria.
      \fi
    }
\end{figure}

\ifthesis
In order to convert the MEDLINE raw text into a form that enables
the deep-learning of scientific hypotheses, we propose a significant
pre-processing pipeline. In short, we begin by downloading relevant data from
Medline and SemMedDB, then extract all relevant information per-sentence to
formulate a semantic graph, following the schema in
Figure~\ref{fig:graph_schema}. From there, we embed the entire network using
PTBG~\cite{lerer2019biggraph}. We then formulate training-set SemRep
predicates~\cite{arnold2015semrep} as sets of node embeddings for our proposed
\textit{transformer encoder} neural-network model, depicted in
Figure~\ref{fig:hypothesis_prediction_model}. 
This pipeline overall is depicted in Figure~\ref{fig:system_diagram}, and is expound below.
\else
    We propose a significant data processing pipeline (Fig.~\ref{fig:system_diagram}), to convert raw-text sources into a semantic graph (Fig.~\ref{fig:graph_schema}). An embedding of this graph enables our learned ranking criteria.
\fi

\noindent{\bf Text Pre-Processing.} We begin with raw MEDLINE XML files
\footnote{At the time of writing, the bulk release at the
end of 2019 contained 1,014 files, containing nearly 30-million documents}.
\ifthesis
Each
can be independently processed for a majority of the \pymoliere{}
distributed data processing operations.
\fi
We attempt to extract the paper id (PMID),
version, title, abstract text, date of first occurrence, keywords, and publication
language.
Next, we filter out non-English documents.
\ifthesis
About 15\% of MEDLINE
documents were not originally published in English, and of those translated a
vast majority contain only a title.
\fi
\emph{In order to validate our system, we
additionally discard any document that is dated after January 1$^{\text{st}}$,
2015.}

We split the text of each abstract
into sentences.  For
each sentence, we identify parts-of-speech, dependency tags, and named entities using ScispaCy~\cite{neumann2019scispacy}.
\ifthesis
We ran
into performance challenges when parsing longer texts. Therefore, we first
perform sentence splitting through a rules-based system provided by the Natural
Language Toolkit (NLTK)~\cite{loper2002nltk}.
\fi
The result of this process is a
record per-sentence, including the title, that contains all metadata associated
with the original abstract, as well as all algorithmically identified annotations.

Using the lemma information of each sentence, we perform $n$-gram mining in
order to identify common phrases that may not have been picked up by entity
detection.
\ifthesis
In our prior work~\cite{sybrandt2017}, we
leveraged a then-contemporary phrase mining tool ToPMine~\cite{elkishky2015scalable}
to extract similar phrases. However, in our new distributed
paradigm, we found it to be more efficient and produce more useful results to
devise a simple rules-based system built on top of ScispaCy lemma information.
\fi
First, we provide a set of
part-of-speech tags we mark as ``interesting'' from the perspective of $n$-gram
mining.
These are: nouns, verbs, adjectives, proper nouns, adverbs, interjections, and ``other.''
We additionally supply a short stopword list, and assert that stop words are
uninteresting.
Then, for each sentence, we produce the set of $n$-grams of length two-to-four
that both start and end with an interesting lemma.
We record any $n$-gram that achieves an overall support of at least 100.
However, we find it necessary to introduce an approximation factor, that an
$n$-gram must have a minimum support of five within a datafile for those
occurrences to count.

\noindent{\bf Semantic Graph Construction.} After splitting sentences, while
simultaneously identifying lemmas, entities and $n$-grams, we can begin
constructing the semantic graph. 
\ifthesis
The semantic graph, as a whole, contains all
textual and biomedical entities within MEDLINE, and follows the schema depicted
in Figure~\ref{fig:graph_schema}.
\fi
We begin this process by creating edges
between similar sentences.  The simplest edge we add is that between two
adjacent sentences from the same abstract. For instance, sentence $i$ in
abstract $A$ will produce edges to $A_{i-1}$ and $A_{i+1}$, with the paper title
serving as $A_0$.

To capture edges between similar sentences in different
abstracts, we compute an approximate-nearest-neighbors network on the set of
sentence embeddings. We derive these embeddings from the average of the final hidden layer of the
SciBert
\footnote{We specifically use the pre-trained
``scibert-scivocab-uncased'' model, which was trained on over 1.14-million
full-text papers.}
NLP model for scientific
text~\cite{beltagy2019scibert}.
This 768-dimensional embedding captures context-sensitive content regarding each word in each sentence.

However, we have over 155-million sentences in the 2015 validation instance of
\pymoliere{}, which makes performing a nearest-neighbors search per-sentence
(typically $\mathcal{O}(n^2d)$) computationally difficult. Therefore, we
leverage FAISS to perform dimensionality reduction, as well as
approximate-nearest neighbors, in a distributed setting.
First, we collect a one-percent sample of all embeddings
on a single machine, wherein we perform product
quantization (PQ)~\cite{jegou2010product}. This technique learns an efficient bit
representation of each embedding. 
\ifthesis
We select parameters for PQ such that each
dimension of the input embedding receives a unique bit in the output code.
\fi
We use 96-quantizers, and each considers a disjoint an
8-dimensional chunk of the 768-dimensional SciBert embeddings. Each quantizer
then learns to map its input real-valued chunk into output 8-bit codes, such
that similar input chunks receive output codes with low hamming distance. 
\ifthesis
This
technique reduces size of SciBert embeddings by a factor of 32.
\fi

Still using the $1\%$ sample on one machine, FAISS performs $k$-Means over PQ
codes in order to partition the reduced space into self-similar buckets. By
storing the centroid of each bucket, we can later select a relevant sub-space
pertaining to each input query, dramatically reducing the search space.  We
select $2048$ partitions to divide the space, and when performing a query, each
input embedding is compared to all embeddings residing in the 16 most-similar
buckets.

Once the PQ quantizers and $k$-means buckets are determined, the initial
parameters are distributed to each machine in the cluster. Every sentence can be
added to the FAISS nearest-neighbors index structure in parallel, and then the
reduced codes and buckets can be merged in-memory on one machine. We again
distributed the nearest-neighbors index, now containing all 155-million sentence
codes, to each machine in the cluster. In parallel, these machines can identify
relevant buckets per-point, and record their 25 approximate nearest-neighbors.
If we have $m$ machines, each with $p$ cores, and search $q=16$ of the $b=2048$
buckets-per-query, we reduce complexity for identifying all nearest-neighbors
from $\mathcal{O}(dn^2)$ to $\mathcal{O}\left(\nicefrac{qdn^2}{32bpm}\right)$.

We additionally add simpler
sentence-occurrence edges for lemmas, $n$-grams and entities.
In each case, we 
produce an edge between $s$ and $x$ provided that lemma, entity, $n$-gram, or
metadata-keyword~$x$ occurs in sentence~$s$.
The last node type is SemRep predicates~\cite{arnold2015semrep}.  Each
has associated metadata, such as the sentence in which it occurred,
its raw text, and its relevant UMLS coded terms. For each unique
subject-verb-object triple, we create a node in the semantic graph. We then
create edges from that node to each relevant sentence, keyword, lemma, entity,
and $n$-gram.
Our overall graph consists of \emph{184-million nodes and 12.3-billion edges}.
\ifthesis
We store this representation of our network first in a series of
Tab-Separated-Value (TSV) files, and provide export utilities to compile this
information into both MongoDB and Sqlite3 databases.
\fi

\noindent{\bf Graph Embedding.}
We
utilize the PyTorch-BigGraph (PTBG) embedding utility to perform a distributed
embedding of the entire network~\cite{lerer2019biggraph}.
\ifthesis
This requires an
expensive index operation from our distributed edge-list structure to
partitioned nodes and bucketed edges expected by PTBG. For this, we provide an
efficient and optimized C++ utility to perform this index in parallel.
\fi
PTBG
learns typed embeddings, and we define node types corresponding to each presented in our semantic graph schema.
Each undirected edge
in our graph schema is also coded as two directional edges of types
$x\rightarrow y$ and $y\rightarrow x$.

\ifthesis
While there are many different configurations possible for PTBG, we explored a
subset to settle on a balance between computational efficiency and embedding
quality.  We partition the semantic graph into 100 roughly even sized partitions
per-type by hashing each node's id string. This partitioning results in 10,000
edge buckets, one for each ordered pair of partitions, which easily fit in one
machine's memory.
\fi
We explore two different embedding dimensionalities: 256 and
512. When computing both embeddings, we specify for edges to be encoded via the
dot-product of nodes, and for relationship types to be encoded using a learned
translation per-type. We generate a total of 100 negative samples per edge, 50
chosen from nodes within each batch, and 50 chosen from nodes within the
corresponding partitions. Dot products between embeddings are learned using the
supplied softmax loss, with the first dimension of every embedding acting as a
bias unit.

Formally, if an edge $ij$ exists between nodes $i$ and $j$ of types $t_i$ and
$t_j$ respectively, then we learn an embedding function $e(\cdot)$ that is used
to create a score for $ij$ by projecting each node into $\mathbb{R}^N$ where $N$
is a predetermined embedding dimensionality. In our experiments we consider
$N=256$ and $512$.  This embedding function uses the typed translation vector
$T^{(t_it_j)}\in \mathbb{R}^N$ that is shared for all edges of the same type as
$ij$. This score is defined as:
\begin{equation}
  s(ij) = e(i)_1 + e(j)_1 + T_1^{(t_it_j)} +
  \sum_{k=2}^N e(i)_k \left(e(j)_k + T_k^{(t_it_j)}\right)
\end{equation}

Then, for each edge $ij$, we generate 100 negative samples in the form
$x_n^{(ij)}y_n^{(ij)}$.  Their scores are compared to that of the positive
sample using the following loss function, which indicates the component of
overall loss corresponding to edge $ij$:
\begin{equation}
  \text{GraphLoss}_{ij}=
  -s(ij)
  +\log\sum_{n=0}^{100}
    \text{exp}\left(
      s\left(
        x_n^{(ij)}y_n^{(ij)}
      \right)
    \right)
\end{equation}

\ifthesis
\noindent \emph{Deployment technical note}: When optimizing our semantic graph embedding, we find that maximal performance
is achieved using a compute cluster of twenty twenty-four-core machines. Within
the 72h time restriction of the Palmetto super computing cluster, we have
enough time to see every edge in the graph 10 times, in the case of the
256-dim embedding, and 5 times in the case of the 512-dim
embedding. Once complete, we are ready to begin training the \pymoliere{}
deep learning hypothesis generation model.
\fi

\ifthesis
\begin{table}[h]
\centering
\begin{tabular}{lr}
    \pymoliere-512 Parameters & 1,313,280 \\
    \pymoliere-512 Embeddings & 10,115,707,904 \\
    \pymoliere-256 Parameters & 328, 960 \\
    \pymoliere-256 Embeddings & 5,057,853,952 \\
\end{tabular}
\caption{Model Size. Because embeddings are trained separately from the hypothesis prediction model, both numbers are listed. Embedding numbers correspond to the amount of floating-point values associated with predicate and coded-term embeddings needed to use the model.}
\label{tab:model_size}
\end{table}

\fi

\noindent{\bf Training Data.} 
In order to learn what makes a plausible biomedical connection, we collect the set of published connections present in our pre-2015 training set.
For this, we turn to the Semantic Medical Database (SemMedDB), which contains over 19-million pre-2015 SemRep~\cite{arnold2015semrep} predicates parsed from all of MEDLINE.
A SemRep predicate is a published subject-verb-object triple that is identified algorithmically.
In lieu of a true data set of attempted hypotheses, we can train our model on these published connections.
However, this approach comes with some drawbacks.
Firstly, SemRep predicates are defined on the set of UMLS terms, which will restrict
our system to only those entities that have been coded. This limitation is
acceptable given size size of UMLS, and presence of existing benchmarks defined
among UMLS terms~\cite{sybrandt2018a}. Secondly, the predicate set is noisy, and may contain entries that are incorrect or obsolete, as well as algorithmically introduced inaccuracies.
However, we find at scale that these sources of noise do not overwhelm the useful signal present within SemMedDB.

\section{Ranking Plausible Connections}
\label{sec:predicate_model}

We train a model to rank published SemRep~\cite{arnold2015semrep} predicates above noisy negative samples using the transformer architecture~\cite{vaswani2017attention}.
To do so we first formulate a predicate with subject $\alpha$ and object $\beta$ for input into the model.
Those predicates that are collected from SemRep are ``positive samples'' (PS).
The function $\Gamma(\cdot)$ indicates the set of neighbor predicates that include a term as either a subject or object.
We represent the $\alpha\beta$ predicate as a set with elements that include both terms, as well as a fixed-size sample with-replacement of size $s=15$ of each node's non-shared predicates:
\begin{equation}
  \begin{split}
    \text{PS}_{\alpha\beta} = \left\{
      \alpha,
      \beta,
      \gamma_1^{(\alpha)},\dots,\gamma_s^{(\alpha)},
      \gamma_1^{(\beta)},\dots,\gamma_s^{(\beta)}
    \right\} \\
    \text{where } \gamma_i^{(\alpha)} \sim \{\Gamma(\alpha)-\Gamma(\beta)\}
    \text{, and } \gamma_i^{(\beta)} \sim \{\Gamma(\beta)-\Gamma(\alpha)\}
  \end{split}
\end{equation}

\noindent{\bf Negative Samples}
We cannot learn to rank positive training examples in isolation. Instead, we
first generate negative samples to accompany each published
predicate. This include two types of samples: scrambles and swaps.
Both are necessary, as we find during training that the easier-to-distinguish scrambles aid early convergence, while the swaps require the model to understand the biomedical concepts encoded by the semantic graph embedding.

The negative scramble (NScr) selects two arbitrary terms $x$ and $y$, as well as $2s$ arbitrary predicates from the set of training data.
While we enforce that $x$ and $y$ do not share a predicate, we do not enforce any relationship between the sampled predicates and these terms.
Therefore these samples are easy to distinguish from positive examples.
If $T$ denote all positive-set terms, and $P$ denotes all predicates, then a negative scramble associated with positive sample $\alpha\beta$ is notated as:
\begin{equation}
\begin{split}
    \text{NScr}_{\alpha\beta} = \left\{x, y, \gamma_1,\dots,\gamma_{2s}\right\}\\
    \text{where } x,y \sim T
    \text{, and } \gamma_i \sim P \\
    \text{s.t. } \Gamma(x)\cap\Gamma(y)=\emptyset
\end{split}
\end{equation}

The negative swap (NSwp) selects two arbitrary terms, but samples the associated predicates in the same manner as the positive sample. 
Therefore, the observed term-predicate relationship will be the same for each half of this negative sample ($\alpha$ and $\gamma_i^{(\alpha)}$).
This sample requires the model to learn that some $\alpha\beta$ pairs should not go together, and this will require an understanding of the relationships between biomedical terms.
A negative scramble associated with $\alpha\beta$ is notated as:
\begin{equation}
\begin{split}
    \text{NSwp}_{\alpha\beta} = \left\{
      x,
      y,
      \gamma_1^{(x)},\dots,\gamma_s^{(x)},
      \gamma_1^{(y)},\dots,\gamma_s^{(y)}
    \right\} \\
    \text{where } \gamma_i^{(x)} \sim \{\Gamma(x)-\Gamma(y)\},
    \text{, and } \gamma_i^{(y)} \sim \{\Gamma(y)-\Gamma(x)\}\\
    \text{s.t. }\Gamma(x)\cap\Gamma(y)=\emptyset
\end{split}
\end{equation}

\noindent{\bf Objective.}
We minimize the margin ranking loss between each positive sample and all associated negative samples.
The contribution of positive sample $\alpha\beta$ to the overall loss is defined as:
\begin{equation}
\begin{split}
\mathcal{L}(\alpha,\beta) 
    = \sum_{i=0}^{n} L\left(\text{PS}_{\alpha\beta}, \text{Nscr}_{\alpha\beta}^{(i)}\right)
    + \sum_{j=0}^{n'} L\left(\text{PS}_{\alpha\beta}, \text{Nswp}_{\alpha\beta}^{(j)}\right) \\
\text{where } L(p, n)
    = \text{max}\left(0, m - \mathcal{H}(p) + \mathcal{H}(n)\right)
\end{split}
\end{equation}

Here $n=10$ denotes the number of negative scrambles, $n'=30$ is the number of negative swaps, $m=0.1$ is the desired margin between positive and negative samples, and $\mathcal{H}$ is the learned function that produces a ranking criteria given two terms and a sample of predicates.

\noindent{\bf Model.}
Using the transformer encoder summarized in Section~\ref{sec:background},
as well as the semantic graph embedding, we construct our model. 
If $e(x)$ represents the semantic graph embedding of $x$, FF represents a feed-forward layer, and $\mathcal{E}$ represents an encoder layer, then our model $\mathcal{H}$ is defined as:
\begin{equation}
\begin{aligned}
    \mathcal{H}(X) &= \text{sigmoid}(\mathcal{M}W) \\
    \mathcal{M} &= \frac{1}{|X|} \sum_{x_i \in X} E_N(FF(e(x_i))) \\
    E_{i+1}(x) &= \mathcal{E}(E_i(x))
    \text{, and } E_0(x) = x
\end{aligned}
\end{equation}

Here $N=4$ represents the number of encoder layers, and $W$ indicates the learned weights associated with the final ranking projection. 
By averaging the transformer output over the input sequence $X$, then projecting that result down to a single real value with $W$, and applying the sigmoid function, we produce an output per-predicate in the unit interval. This function is depicted in Figure~\ref{sec:predicate_model}.
\ifthesis
\noindent \emph{Deployment technical note}: We minimize the ranking loss over
all published predicates using the LAMB optimizer \cite{you2019large}. This
allows us to efficiently train using very large batch sizes, which is necessary
as we leverage 10 NVIDIA V100 GPUs to effectively process 600 positive samples
(and therefore 2,400 total samples) per batch. In terms of hyperparameters, we
select a learning rate of $\eta=0.01$ with a linear warm up of 1,000 batches, a
margin of $m=0.1$, a neighborhood sub-sampling rate of $s=15$, and we perform
cross-validation on a $1\%$ random holdout to provide early stopping and to
select the best model with respect to validation loss. Due to the large size of
training data, one epoch consists of only $10\%$ of the overall training data.
This process is made easier through the helpful Pytorch-Lightning
library~\cite{falcon2019lightning}.
\else
The supplemental information containing training parameters and additional model detail.
\fi

\section{Validation}
\label{sec:validation}

Testing hypothesis generation, in contrast to information retrieval, is
difficult as ultimately these systems are intended to discover information that
is unknown to even those designing them~\cite{yetisgen2008evaluation}. A
thorough evaluation would require a costly process wherein scientists explore
automatically posed hypotheses. Instead, we perform a historical validation, in
a manner similar to that performed in~\cite{sybrandt2018a, sybrandt2018b}.  This
method enables large-scale evaluation of many biomedical subdomains almost
instantly, but cannot truly tell us how our system will perform in a laboratory
environment.
\ifthesis
To attempt to incorporate some expert oversight into the validation
process, we supplement automatic validation with qualitative analysis from a
domains scientist, which follows the older validation process found
in~\cite{sybrandt2017}.  After ensuring the system is capable of uncovering
recent connections from historical data, we begin the much longer process of
testing contemporary ideas system in real-world scenarios, as was pursued by
Moliere~\cite{aksenova2019inhibition}, Watson~\cite{bakkar2018artificial}, and
ARROWSMITH~\cite{swanson1986fish}.
\fi

\noindent{\bf Comparison with Heuristic-Based Ranking.}
We begin by comparing the performance numbers obtained through our proposed learned ranking criteria with other ranking methods posed in~\cite{sybrandt2018a}.
Specifically, the Moliere system presents experimental numbers for various training-data scenarios for the same 2015 temporal holdout as used in this work~\cite{sybrandt2018b}.
For a direct comparison, we use our proposed method to rank the same set of positive and negative validation examples.

\noindent{\bf Comparison by Subdomain Recommendation.}
As mentioned in~\cite{henry2019indirect}, the Moliere validation set has
limitations. We improve this set by expanding both the quantity and diversity of
considered term pairs, as well as evaluating \pymoliere{} through the use of
all-pairs recommendation queries within popular biomedical subdomains.
As a result, this comparison effectively uses subdomain-specific negative examples, which makes for a harder benchmark than that presented in the Moliere work.
It is worth nothing that these all-pairs searches are made possible by the very efficient neural-network inference within \pymoliere{}, and would not be as
computationally efficient in the Moliere shortest-path and topic-modeling approach.

This analysis begins by extracting {\it semantic types}~\cite{semantictypes}, which categorize each UMLS term per-predicate into
one of 134 categories, including ``Lipid,'' ''Plant,'' or ``Enzyme.''
From there, we can group $\alpha\beta$
predicate-term pairs by types $t_\alpha$ and $t_\beta$. We select the twenty predicate
type pairs with the most popularity in the post-2015 dataset, and within each
type we identify the top-100 predicates with the most rapid non-decreasing
growth of popularity determined by the number of abstracts containing each term-pair
per year.  These predicates form the positive class of the validation set. We
form the rest of the subdomain's validation set by recording all possible
undiscovered pairs of type $t_{\alpha}t_\beta$ from among the UMLS terms in the top-100
predicates. We then rank the resulting set by the learned ranking criteria, and
evaluate these results using a range of metrics.

\noindent{\bf Metrics.}
The first metrics we consider are typical for determining a classification
threshold: the area under the receiver-operating-characteristic curve (AUC ROC)
and the area under the precision-recall curve (AUC PR).
\ifthesis
An AUC ROC of 0.5
indicates that the ranking criteria randomly orders the published term pairs
relative to the undiscovered, while an area of 1 indicates that all published
term pairs occur first in the ranking. Similarly, the PR curve determines how
varying levels of precision could be achieved while still retrieving a certain
amount of published connections. An AUC PR closer to 1 again indicates that all
published term pairs occur at the start of the list, and a PR closer to 0
indicates that they occur towards the end. However unpublished predicates
occurring to the start of the list typically have a larger negative impact in
the score.
\fi
We additionally provide recommendation system metrics, such as top-$k$ precision
(P.@$k$), average precision (AP.@$k$), and overall reciprocal rank (RR). Top-$k$
precision is simply the number of published term-pairs appearing in the first $k$
elements of the ranked list, divided by $k$. Top-$k$ \textit{average} precision
weights each published result by its location in the front of the ranked list.
The reciprocal rank is the inverse of the rank of the first published term pair.

The above recommender system metrics all consider the single many-to-many query
within a biomedical subdomain. However, this same result can be interpreted as a
set of one-to-many recommendation queries. Doing so enables us to compute the
mean average precision(MAP.@$k$), and mean-reciprocal rank(MRR.@$k$) for the set
of recommendations.
A high MRR within a domain indicates that the researcher should expect to see a useful result within the first few results. A high MAP indicates that out of the top $k$ results, more of them are useful.
These metrics, taken together, should influence biomedical researchers when exploring the results of a one-to-many query.

\ifthesis
While all of the above metrics quantify the performance of the \pymoliere{}
learned ranking criteria, it is also important to provide interpretable results
to biomedical researchers. For this reason we also perform Moliere-style
shortest-path and topic-model queries on the \pymoliere{} semantic graph. Our
newly optimized framework enables us to perform one query on a single thread in
a few minutes, which can allow a medical researcher to explore a subset
of recommendations output by the deep learning model. We visualize these topic
model outputs and provide them to domain scientists in order to give feedback on
their predictive power for recent findings. One such finding, the relationship
between HIV-associated Neurodegenerative Disease, which was found by Moliere in
2019~\cite{aksenova2019inhibition}, is among this qualitative study.
\fi
\section{Results}
\label{sec:results}

\begin{table*}
{\small
\begin{tabular}{c|rr|rr|r|rr|rr|rr|rr}
\toprule

            & \multicolumn{2}{c|}{Training}
            & \multicolumn{2}{c|}{AUC}
            & %RR
            & \multicolumn{2}{|c}{P.@}
            & \multicolumn{2}{|c}{AP.@}
            & \multicolumn{2}{|c}{MAP.@}
            & \multicolumn{2}{|c}{MRR.@}
            \\
       Type &       \% &   Rank &   PR &  ROC &   RR &    10 &    100 &     10 &     100 &      10 &      100 &      10 &      100 \\
\midrule
 gngm, celf &     0.29 &     74 & 0.44 & 0.62 & 1.00 &  0.50 &   0.47 &   0.83 &    0.54 &    0.57 &     0.56 &    0.61 &     0.61 \\
 gngm, neop &     0.35 &     61 & 0.34 & 0.65 & 0.50 &  0.50 &   0.43 &   0.54 &    0.47 &    0.46 &     0.41 &    0.52 &     0.52 \\
 aapp, neop &     0.35 &     62 & 0.20 & 0.62 & 0.33 &  0.30 &   0.26 &   0.34 &    0.28 &    0.40 &     0.35 &    0.46 &     0.47 \\
 gngm, cell &     0.43 &     42 & 0.19 & 0.72 & 0.25 &  0.30 &   0.17 &   0.27 &    0.21 &    0.35 &     0.32 &    0.38 &     0.38 \\
 aapp, cell &     0.67 &     26 & 0.19 & 0.63 & 0.50 &  0.20 &   0.17 &   0.36 &    0.21 &    0.34 &     0.33 &    0.37 &     0.38 \\
 aapp, gngm &     1.05 &     13 & 0.17 & 0.68 & 1.00 &  0.50 &   0.22 &   0.61 &    0.31 &    0.36 &     0.27 &    0.39 &     0.40 \\
 cell, aapp &     1.59 &      4 & 0.17 & 0.67 & 0.14 &  0.10 &   0.19 &   0.14 &    0.18 &    0.35 &     0.32 &    0.40 &     0.41 \\
 gngm, gngm &     0.50 &     37 & 0.17 & 0.66 & 1.00 &  0.40 &   0.20 &   0.77 &    0.37 &    0.31 &     0.26 &    0.33 &     0.34 \\
 orch, gngm &     0.41 &     49 & 0.16 & 0.69 & 0.05 &  0.00 &   0.22 &   0.00 &    0.21 &    0.33 &     0.27 &    0.34 &     0.36 \\
 aapp, dsyn &     0.67 &     25 & 0.15 & 0.69 & 0.33 &  0.20 &   0.24 &   0.28 &    0.22 &    0.34 &     0.27 &    0.37 &     0.38 \\
 gngm, dsyn &     0.21 &     97 & 0.15 & 0.71 & 0.50 &  0.40 &   0.24 &   0.59 &    0.32 &    0.29 &     0.23 &    0.30 &     0.31 \\
 bpoc, aapp &     1.06 &     12 & 0.14 & 0.67 & 1.00 &  0.20 &   0.18 &   0.70 &    0.28 &    0.35 &     0.30 &    0.38 &     0.39 \\
 bacs, gngm &     0.29 &     73 & 0.12 & 0.67 & 0.33 &  0.10 &   0.14 &   0.33 &    0.19 &    0.26 &     0.24 &    0.29 &     0.30 \\
 bacs, aapp &     0.73 &     22 & 0.12 & 0.68 & 0.17 &  0.30 &   0.14 &   0.28 &    0.18 &    0.27 &     0.24 &    0.30 &     0.32 \\
 dsyn, humn &     7.02 &      1 & 0.11 & 0.64 & 0.05 &  0.00 &   0.10 &   0.00 &    0.10 &    0.27 &     0.25 &    0.29 &     0.31 \\
 aapp, aapp &     1.57 &      5 & 0.11 & 0.69 & 1.00 &  0.20 &   0.11 &   0.67 &    0.25 &    0.28 &     0.24 &    0.32 &     0.33 \\
 gngm, aapp &     0.40 &     52 & 0.11 & 0.71 & 1.00 &  0.20 &   0.11 &   0.61 &    0.22 &    0.23 &     0.21 &    0.25 &     0.26 \\
 phsu, dsyn &     0.76 &     20 & 0.10 & 0.61 & 0.04 &  0.00 &   0.14 &   0.00 &    0.11 &    0.27 &     0.20 &    0.30 &     0.31 \\
 dsyn, dsyn &     1.35 &      6 & 0.09 & 0.62 & 0.17 &  0.20 &   0.12 &   0.19 &    0.15 &    0.22 &     0.18 &    0.25 &     0.27 \\
 topp, dsyn &     1.19 &      9 & 0.09 & 0.64 & 0.10 &  0.10 &   0.17 &   0.10 &    0.12 &    0.28 &     0.22 &    0.30 &     0.31 \\
\bottomrule
\end{tabular}
\caption{
  \label{tab:pymoliere_512_per_type}
  \pymoliere-512. Above are hypothesis prediction results on biomedical
  sub-domains. Indicated along with performance numbers are the percentage of
  training data (pre-2015 predates) as well as the training-data popularity rank
  out of 6396, with 1 being most popular.
  Metrics described in detail in Section~\ref{sec:validation}.
 }
 }
\end{table*}

We compare the performance of \pymoliere{} against Moliere, as
presented in~\cite{sybrandt2018b}. In that work, multiple trained instances of Moliere
rank a benchmark set of positive and negative potential connections using a
range of criteria defined in~\cite{sybrandt2018a}.  These Moliere instances each
use different datasets published prior to 2015 in order to perform hypothesis
queries, of which we focus on two: all of MEDLINE (Moliere: MEDLINE), and all of
PubMedCentral (Moliere: Full Text). The former instance represents a system
trained on the same raw data as the \pymoliere{} system presented here, while
the latter represents a system trained on all publicly available full-text
papers provided by the NLM released in the same date range.

The prior work establishes that the Moliere topic-modeling approach is improved
by the additional information made available by full-text papers, but at a
overwhelming 45x runtime penalty. These  quality results are reproduced in
Table~\ref{tab:2015_benchmark}, and we include additional results for the
\pymoliere{} system when evaluated on only abstracts, and exactly the same set
of predicates. We observe that the \pymoliere{} system, when trained with
512-dimensional graph embeddings, improves upon Moliere: Medline by 25\% and
Moliere: Full Text by 13\%. Importantly, this increase in quality comes at an
overwhelming \textit{decrease} in runtime, with the wall time per-query dropping
from minutes to milliseconds, due to the introduction of the deep-learning
approach. 
\begin{table}[]
    \centering
    \begin{tabular}{lrr}
      \toprule
      System Instance   & ROC AUC & PR AUC \\
      \midrule
      Moliere: Medline & 0.718   & 0.820  \\
      Moliere: Full Text & 0.795   & 0.778  \\
      \pymoliere-256    & 0.826   & 0.895  \\
      \pymoliere-512    & {\bf0.901}   & {\bf0.936} \\
      \bottomrule
    \end{tabular}
    \caption{
      \label{tab:2015_benchmark}
      Benchmark comparison between Moliere and \pymoliere{} on the same
      benchmark.
    }
\end{table}

\ifthesis
Figures~\ref{fig:validation_2015_roc} and~\ref{fig:validation_2015_pr}
depict ROC and PR curves for the top-performing \pymoliere{} model.
\begin{figure}
    \centering
    \includegraphics[width=0.9\linewidth]{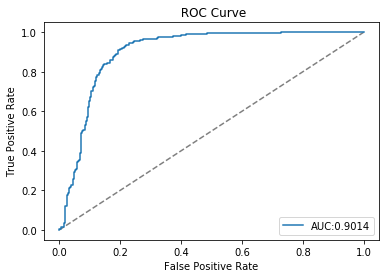}
    \caption{Validation Benchmark 2015 ROC}
    \label{fig:validation_2015_roc}
\end{figure}

\begin{figure}
    \centering
    \includegraphics[width=0.9\linewidth]{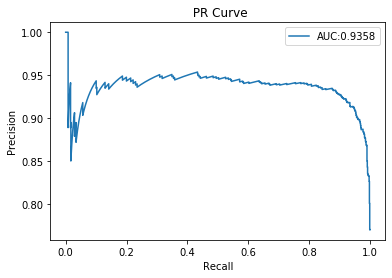}
    \caption{Validation Benchmark 2015 PR}
    \label{fig:validation_2015_pr}
\end{figure}
\fi

\ifthesis
To further study the performance differences between the Moliere and
\pymoliere{} systems, we quantify the correlations between their different
ranking criteria.  We depict these correlations with Moliere: Medline and
Moliere: Full Text in Figures~\ref{fig:corr_pymoliere_abstract}
and~\ref{fig:corr_pymoliere_fulltext} respectively. Each point in these scatter
plots indicate a hypothesis, which is colored green if it was published
following 2015, and red if it was negatively sampled. The scale for both scatter
plots is determined by the intervals spanned by each system's ranking criteria.
We observe that there is very little correlation between these scores, and that
the separation between positive and negative samples is clearly seen in the
\pymoliere{} ranking, and muddied in the Moliere rankings. As a result, we do not
believe there would be a substantial benefit in creating an ensemble method to
combine the deep-learning and classical ranking methods.

\begin{figure}
    \centering
    \includegraphics[width=0.9\linewidth]{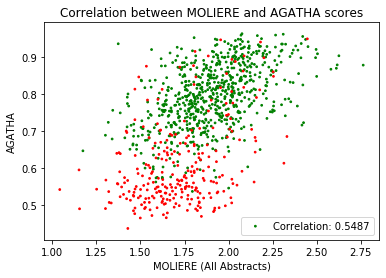}
    \caption{Comparisons between \pymoliere-512 scores and Moliere: Medline scores on the 2015 benchmark.}
    \label{fig:corr_pymoliere_abstract}
\end{figure}

\begin{figure}
    \centering
    \includegraphics[width=0.9\linewidth]{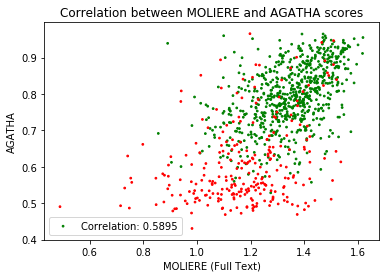}
    \caption{Comparisons between \pymoliere-512 scores and Moliere: Full Text scores on the 2015 benchmark.}
    \label{fig:corr_pymoliere_fulltext}
\end{figure}
\fi

To extend the validation beyond the above results, provided that we can now
generate thousands of hypothesis per-minute, we explore the capacity of our
deep-learning ranking criteria to perform hypothesis recommendation within
various many-to-many queries across different biomedical sub-domains. These
results, displayed in Table~\ref{tab:pymoliere_512_per_type}, list the 20
predicate types with the most popularity following 2015. Due to space
limitations, we present predicate types using NLM semantic type
codes~\cite{semantictypes}. All numbers are reported from the \pymoliere-512
model.

We observe that the (Gene)$\rightarrow$(Cell Function)(gngm, celf) predicate
type, is the easiest predicate type for \pymoliere-512 to recommend, even though
connections of this type only account for 0.29\% of the training data. Of the
top-10 recommendations the highest ranked is a valid connection and half are
valuable.  When performing a one-to-many query within this type of connection,
we observe 85\% of all top-10 suggestions to be useful on average, and that a
useful result occurs typically within the first two recommendations.  We see
similar performance in the (Gene)$\rightarrow$(Neoplastic Process) (gngm, neop)
and (Amino Acid, Peptide, or Protein)$\rightarrow$(Neoplastic Process) (aapp,
neop) sub-domains. Interestingly, there appears to be little correlation between
the popularity of a predicate type in the training data and the quality of the
resulting recommendations. This result enforces the idea of \pymoliere{} as a
\textit{general-purpose} biomedical hypothesis generation system.

\sloppy
Of the 20-most-popular predicate subdomains considered, {\pymoliere{}-512} has the
most difficulty with the (Therapeutic or Preventive
Procedure)$\rightarrow$(Disease or Syndrome)(topp, dsyn). In this subdomain,
the the highest ranked positive predicate is ranked tenth, and only twelve of the
top-100 suggestions are useful. Still, in a one-to-many query, we expect about
one-in-ten recommended predicates to be useful, and for the top-3 predicates to
contain a useful result. While the lower-performing subdomains are significantly
harder for \pymoliere-512 than the top few, we note that even a low-precision
tool can be useful for aiding the biomedical discovery process. Furthermore,
these difficult subdomains are still ranked significantly better than random
chance, and even better than many of the classical ranking measures presented
in~\cite{sybrandt2018a}. Using this information, future work may wish to
fine-tune the \pymoliere{} method to a specific subdomain for improved
performance.

\section{Lessons Learned and Open Problems}
\label{sec:lessons_learned}

{\bf Result Interpretability}. While deep-learning models are notoriously hard
for human decision makers to interpret, we find that biomedical researchers
still need to understand how a result was produced in order to act on model
predictions. However, we cannot leave the entire analysis up for human
judgement, as this drastically reduces the benefits of ``automatic'' hypothesis
generation. To walk the narrow edge between these conflicting objectives, we
implement both an automatic ranking component, as well as a more interpretable
topic-model query system. We find that these tools serve different functions
during different times of the discovery process.

At first, a researcher may be considering a wide range of potential research
directions, such as during the candidate selection phase of the drug-discovery
process. This often requires assembling hundreds (or thousands) of target
ingredients, compounds, genes, or deceases, and determining whether elements of
this large set have a relationship to an item of interest. For instance, when we
evaluated HIV-associated Neurodegenerative Disease, we explored over 40,000
potential human genes~\cite{aksenova2019inhibition}. This component of the
discovery process fits nicely into the deep-learning ranking and recommendation
system proposed here, especially when the target set is so large that a manual
literature review may prove costly.

Once a candidate set of targets has been winnowed from the large target set, the
researcher will prioritize interpretability. However, the candidate set is
typically orders of magnitude smaller than the target set. Therefore, we can
afford to run more costly-yet-interpretable routines, even if these routines do
not provide any form of ``automatic'' analysis. At this stage, we switch from
our deep-learning ranking method to the topic-modeling approach similar to that
presented in Moliere~\cite{sybrandt2017}. This process finds a path within our
semantic network containing the textual information necessary to describe a
potential connection. We present that path along with the set of relevant
sentences, as well as a visualization of the topic model built from those
sentences. Researchers can explore the sets of entities that are frequently
mentioned together in order to expand their mental models of each hypothesis's
quality.

{\bf Datasets and Expandability}. When discussing hypothesis generation systems
with prospective adopters in the biomedical community, we often are asked to
include specific datasets that has domain-relevance to an individual's research
direction. For instance, the set of clinical trials, internal experimental
findings, or a database of chemicals.

\ifthesis
Currently, new network-based datasets can be introduced trivially. After
processing Medline, the network lives as a collection of simple TSV files. This
set can receive new datasets, provided that the new entity types are included in
the following PyTorch-BigGraph configuration. We use this graph-addition
technique to merge SemRep predicate data into the Medline dataset.

New textual datasets can similarly be introduced earlier in the process. Text
records, once formulated into python dictionaries with a particular set of
fields, may be added to the pipeline, participate in the tokenization and
network-construction process, and will eventually be included in topic-model
queries. however, this process requires a minor modification to the existing
data pipeline code.  We are working currently to make this import operation as
simply as the network-addition process described above.

\else
    We designed \pymoliere{} to be easily extendable.
    Domain scientists can easily supply new graph datasets as a collection of TSV files and by making minor changes to the PTBG configuration.
    Furthermore, new textual sources can be merged into the pipeline with straightforward modification to the python data-processing pipeline.
\fi
In contrast to the graph and text sources, it is not currently clear how to
incorporate experimental data into the \pymoliere{} system. This challenge
arises from the many forms experimental data can take.  In the case where an
experiment can be reformulated as a network, such as converting the
gene-expression matrix into a gene-to-gene network, these results can trivially
be introduced as new edges. Other experimental results, such as many clinical
trials, include a thorough summary of that trial's findings. These may be
introduced as a combination of textual and graph-based sources, including both
the description text, as well as any links to known publications that reference
the trial. Importantly, we do not find a ``one size fits all'' solution for
experimental data, and more work should explore the costs and benefit associated
with various datasets.

\section{Related Work}
\label{sec:related_work}

Foster et al.~\cite{foster2015tradition} identify a series of common successful research strategies often used by scientists. 
In doing so they demonstrate that high-risk and innovative strategies are uncommon among the scientific community in general.
It follows that the field of hypotheses generation obeys similar rules.
Many systems have found success using algorithmic techniques that approximate these common research strategies by studying term co-occurrences~\cite{jelier2008anni, hristovski2006exploiting, weeber2000text}, or predicting links with a graph of biomedical entities~\cite{pusala2017supervised, eronen2012biomine}.
While the Foster's model of research strategies has proven to be useful, 
the mechanisms involved in complex scientific discoveries remain unexplored.

Unsurprisingly, we find that hypothesis generation systems utilize algorithmic techniques in a range of complexity that is analogous to these human research strategies.
The first hypothesis generation system, ARROWSMITH, presents the ABC model of automatic discovery~\cite{swanson1986fish}. This technique identifies a list of terms that are anticipated to help explain a connection between two terms of interest. This basic algorithm remains in some modern systems, such as~\cite{kim2019context}.
However, ABC-based techniques have significant limitations~\cite{smalheiser2012literature},
including their similarity metrics defined on heuristically determined term lists, as well as their reliance on manual validation processes.
As a result, ABC systems are know to be biased towards finding incremental discoveries~\cite{kostoff2009literature}.

A completely different strategy of performing LBD is proposed by Spangler et al. in ~\cite{spangler2014automated}.
To explore the p53 kinase, the authors use neighborhood graphs constructed from 
entity co-occurrence rates.
The approach relies on domain experts and requires manual oversight to provide MEDLINE search queries, and to prune redundant terms, but produces promising results.
In~\cite{choi2018literature}
the authors demonstrate that this technique can identify kinase NEK2 as an inhibitor of p53, and in~\cite{bakkar2018artificial} a similar 
scientist-in-the-loop technique identifies
a number of RNA-binding proteins associated with ALS.

A significant step beyond ABC and human-assisted techniques is to incorporate a domain specific datasets. Bipartite graphs, such as the gene-disease~\cite{liu2014diseaseconnect} or the term-document~\cite{gopalakrishnan2018towards} networks, are frequent choices.
These  systems usually aim to perform a number of graph traversals between node-pairs in order to rank the most viable options. 
However, the number of generated paths may be prohibitively large, which reduces ranking quality~\cite{gopalakrishnan2016generating}
To address this problem, Gopalakrishnan proposes two-stage filtering through a "single-class classifier" which is able to prune up to 90{\%} hypotheses prior to the ranking scheme~\cite{gopalakrishnan2018towards}

One recent approach is to use deep learning models to help extract viable biomedical hypotheses. Sang et al.~\cite{sang2018gredel} describe GrEDeL, a way to generate new hypotheses using knowledge graphs obtained from predicate triples in the form of “subject, verb, object”. This approach finds all possible paths between a given drug and decease, provided those paths include a particular target entity. Then these paths are evaluated using a LSTM model that captures features related to drug-disease associations.
While the GrEDeL system is successful at identifying some novel drug-disease relationships, this approach has some important trade-offs: 
(1) Their proposed model is trained using SemRep graph traversals as a sequence, which the authors note is a highly noisy dataset. Furthermore, multiple redundant and similar paths exist within their dataset, which decrease the quality of their validation holdout set. The \pymoliere{} system overcomes this limitation by leveraging node neighborhoods in place of paths.
(2) Their knowledge graph is constructed exclusively from predicates mined from MEDLINE abstracts using SemRep. This process affects the model quality significantly and, being the only resource of knowledge, it requires careful manual filtering of false positive and isolated predicates. 
(3) The GrEDeL LSTM model is trained to only discover drug-disease associations, and does not generalize to other biomedical subdomains.
(4) This approach embeds their predicate knowledge graph using the TransE method~\cite{bordes2013translating}, which supposes that relationships can be modeled as direct linear transformations. When using the large number of relationship types present in SemRep, this assumption greatly reduces the useful variance in the resulting node embeddings.
\section{Conclusions}
\label{sec:conclusion}

This work presents \pymoliere{}, a deep-learning biomedical hypothesis
generation system, which can accelerate discovery by learning to detect useful
new research ideas from existing literature. This technique enables domain
scientists to keep pace with the accelerating rate of publications, and to
efficiently extract implicit connections from the breadth of biomedical
research. By constructing a large semantic network, embedding that network, and
then training a transformer-encoder deep-learning model, we can learn a ranking
criteria that prioritizes plausible connections. We validate
this ranking technique by constructing an instance of the \pymoliere{} system
using only data published prior to January 1$^{\text{st}}$ 2015.  This system
then evaluates both a benchmark of predicates established from prior work~\cite{sybrandt2018b}, and
performs recommendation in twenty popular biomedical subdomains. The result is
state-of-the art prediction quality on the 2015 benchmark, as well as strong performance across a range of subdomains.
\ifthesis
In the
case where a simple recommendation is not sufficient, we also implement
topic-model-based interpretability queries, that enable researchers to learn
more about particular connections of interest, after the initial ranking has
limited their field of consideration.
\fi
The \pymoliere{} system is open-source and
written entirely in Python and PyTorch, which enable to be easily used or
adapted anywhere. We release both the 2015 validation system, as well as an
up-to-date 2019 system to accelerate the broader community of biomedical
sciences.

%\begin{acks}
%We would like to thank the CCIT staff who manage the Palmetto Supercomputer at Clemson University, where we ran all of our experiments. This work was additionally supported by NSF \#1633608.
%\end{acks}

\bibliographystyle{plain}
\bibliography{main}

%%\appendix

\ifthesis
\else
\pagebreak
\section*{Reproducability Details}

\noindent{\bf Training Graph Embedding}
When optimizing our semantic graph embedding, we find that maximal performance
is achieved using a compute cluster of twenty twenty-four-core machines. Within
the 72h time restriction of the Palmetto super computing cluster, we have
enough time to see every edge in the graph 10 times, in the case of the
256-dim embedding, and 5 times in the case of the 512-dim
embedding. Once complete, we are ready to begin training the \pymoliere{}
deep learning hypothesis generation model.

\noindent{\bf Training Ranking Model}
We minimize the ranking loss over
all published predicates using the LAMB optimizer \cite{you2019large}. This
allows us to efficiently train using very large batch sizes, which is necessary
as we leverage 10 NVIDIA V100 GPUs to effectively process 600 positive samples
(and therefore 2,400 total samples) per batch. In terms of hyperparameters, we
select a learning rate of $\eta=0.01$ with a linear warm up of 1,000 batches, a
margin of $m=0.1$, a neighborhood sub-sampling rate of $s=15$, and we perform
cross-validation on a $1\%$ random holdout to provide early stopping and to
select the best model with respect to validation loss. Due to the large size of
training data, one epoch consists of only $10\%$ of the overall training data.
This process is made easier through the helpful Pytorch-Lightning
library~\cite{falcon2019lightning}.

\begin{table}[H]
\centering
\begin{tabular}{lrrr}
\toprule
Layer Name & Input Dim. & Output Dim. & Num. Params\\
\midrule
Linear (ReLU) & 512 & 512 & 262656 \\
Enc. M.H.Att.& 512 & 512 & 1050624 \\
Enc. Dropout(0.1) & 512 & 512 & 0 \\
Enc. LayerNorm & 512 & 512 & 1024 \\
Enc. Linear (ReLU) & 512 & 1024 & 524800 \\
Enc. Dropout(0.1) & 512 & 512 & 0 \\
Enc. Linear (ReLU) & 1024 & 512 & 525312\\
Enc. Dropout(0.1) & 512 & 512 & 0 \\
Enc. LayerNorm & 512 & 512 & 1024 \\
\textit{Encoder 2} & 512 & 512 & 2102272\\
\textit{Encoder 3} & 512 & 512 & 2102272\\
\textit{Encoder 4} & 512 & 512 & 2102272\\
Linear (sigmoid) & 512 & 1 & 513 \\
\bottomrule
\end{tabular}
\caption{\label{tab:model_parameters}Layers and parameter counts for the \pymoliere{} transformer model.}
\end{table}
\begin{table}[H]
    \centering
    \begin{tabular}{lr}
    \toprule
        Node Type & Count \\
    \midrule
        Sentence & 140,913,505\\
        Predicate & 19,268,319 \\
        Lemma & 12,718,832 \\
        Entity & 10,240,635 \\
        Coded Term & 488,923 \\
        $n$-Grams & 333,862 \\
    \midrule
        Total Nodes &  183,964,076\\
        Total Edges &  12,362,325,167\\
    \bottomrule
    \end{tabular}
    \caption{Graph Size of 2015 Validation Dataset}
    \label{tab:graph_size}
\end{table}

\fi

\end{document}